\documentclass[conference]{IEEEtran}
\usepackage[noadjust]{cite}
\usepackage{graphicx}
\usepackage{array}
\usepackage{balance}
\usepackage{epsfig}
\usepackage{epstopdf}
\usepackage{subfigure}
\usepackage{multirow}
\usepackage{color}

\newcommand{\softpara}[1]{\smallskip \noindent \underline{#1}}

\usepackage{tikz}
 
\usepackage{silence}
\usepackage{url}

\usepackage{arabtex}
\usepackage{utf8}
\usepackage{tipa}

\begin{document}
\title{Arabic Text Diacritization Using Deep Neural Networks}

\author{\IEEEauthorblockN{Ali Fadel, Ibraheem Tuffaha, Bara' Al-Jawarneh and Mahmoud Al-Ayyoub}
\IEEEauthorblockA{
Jordan University of Science and Technology, Irbid, Jordan
\\ 
\{aliosm1997, bro.t.1996, baraaaljawarneh\}@gmail.com,
maalshbool@just.edu.jo}}

\maketitle
\IEEEpubidadjcol

\begin{abstract}
Diacritization of Arabic text is both an interesting and a challenging problem at the same time with various applications ranging from speech synthesis to helping students learning the Arabic language. Like many other tasks or problems in Arabic language processing, the weak efforts invested into this problem and the lack of available (open-source) resources hinder the progress towards solving this problem. This work provides a critical review for the currently existing systems, measures and resources for Arabic text diacritization. Moreover, it introduces a much-needed free-for-all cleaned dataset that can be easily used to benchmark any work on Arabic diacritization. Extracted from the Tashkeela Corpus, the dataset consists of 55K lines containing about 2.3M words. After constructing the dataset, existing tools and systems are tested on it. The results of the experiments show that the neural Shakkala system significantly outperforms traditional rule-based approaches and other closed-source tools with a Diacritic Error Rate (DER) of 2.88\% compared with 13.78\%, which the best DER for the non-neural approach (obtained by the Mishkal tool).
\end{abstract}


\section{Introduction}
\label{sec:intro}



Arabic is among the most widely spoken languages in the world. It is the native language of hundreds of millions of people and one of the official languages for dozens of countries. The community of Arabic speakers has one of the largest growth rates on the Internet. Thus, the interest in Arabic Natural Language Processing (NLP) has increased over the years. Unfortunately, the work on Arabic NLP is lagging behind the NLP work for other languages, such as English and Chinese, due to many reasons including the poor efforts invested in Arabic NLP and the lack of linguistic resources available for researchers and developers \cite{habash2010introduction,farghaly2009arabic,asa_survey}.

\begin{figure}
    \centering
    \includegraphics[width=0.5\textwidth]{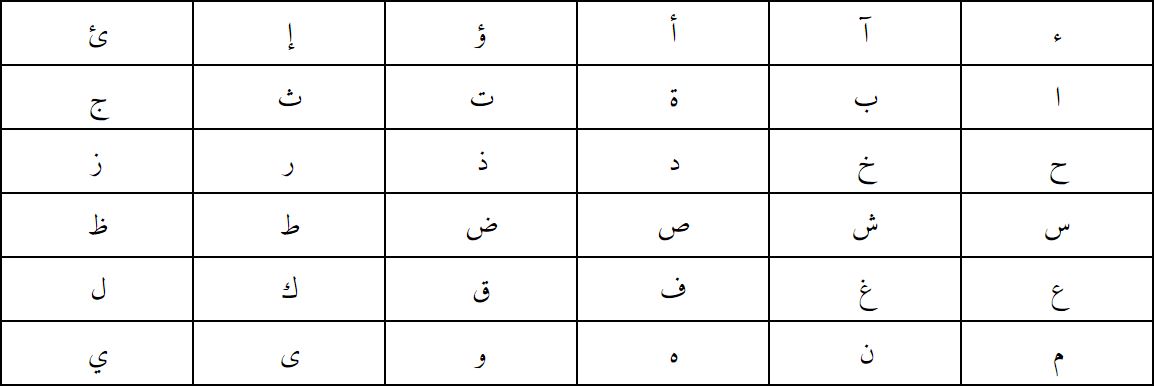}
    \caption{The 36 Arabic letter variants.}
    \label{tab1_1}
\end{figure}

\begin{figure}
    \centering
    \includegraphics[width=0.5\textwidth]{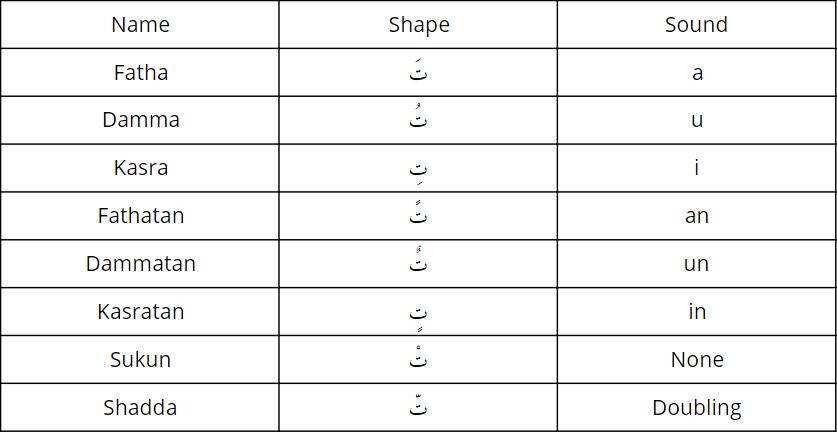}
    \caption{The 8 Arabic basic diacritics on the Teh letter \cite{abandah2015automatic}.}
    \label{tab1_2}
\end{figure}

The Arabic alphabet is the base alphabet used in multiple languages including: Arabic, Persian and Kurdish. The Arabic language has 36 variants (see Figure~\ref{tab1_1}) of the basic 28 letters and eight basic diacritics (see Figure~\ref{tab1_2}) \cite{abandah2015automatic}.

Two of the major aspects differentiating the Arabic language from most other languages are: the right to left (RTL) writing style and the addition of diacritics to each letter as shown in the following example:
\begin{quote}
\centering
\setcode{utf8}\<
اَلْمُسْلِمُ مَنْ سَلِمَ اَلْمُسْلِمُونَ مِنْ لِسَانِهِ وَيَدِهِ
> \\
Buckwalter Transliteration: Aalomusolimu mano salima Aalomusolimuwna mino lisaAnihi wayadihi \\
Translation: A Muslim is the one from whose tongue
and hands the Muslims are safe.
\end{quote}

The diacritics have huge influence on the meaning of the sentences and the diacritization can be affected by the context of the sentence as shown in the following example:
\begin{quote}
\centering
\<
ذهب علي ...
> \\
Buckwalter Transliteration: *hb Ely ... \\
Incomplete sentence without diacritization. \\
\<
ذَهَبَ عَلِيٌّ بَعِيداً
>\\
Buckwalter Transliteration: *ahaba EaliyN$\sim$ baEiydAF \\
Translation: Ali went away. \\
\<
ذَهَبُ عَلِيٍّ كَثِيرٌ
> \\
Buckwalter Transliteration: *ahabu Ealiy$\sim$K kaviyrN \\
Translation: Ali has a lot of gold.
\end{quote}
Given two different diacritizations, the letters
\<
ذهب 
>
represent two different words with different part of speech (POS) tags. As shown in the example above,
\<
ذَهَبَ 
>
``thahaba'' in the first sentence is the verb `went' in English, while
\<
ذَهَبُ
>
``thahabu'' in the second sentence is the noun `gold' in English.

Moreover, Arabic text can be either partially or fully diacritized. This is because in some applications there is no need to diacritize all characters. In addition, Arabic has a diglossic nature manifested through the co-existence of Standard Arabic and Colloquial Arabic. Standard Arabic is also split into two categories: Classical Arabic (CA) and Modern Standard Arabic (MSA). CA is mainly used in the Holy Quran (HQ), old books, old poetry, etc., while MSA is used in news, lectures, letters, formal speeches, etc. The Colloquial Arabic is used in daily life, and diacritization is usually not used for this type of Arabic, unlike CA and MSA \cite{habash2010introduction}.


The Arabic language is one of the most widely used languages. Yet, the attention to using it with proper grammar is very little. Hence, the idea to build an automated diacritization system to help both fluent and non-fluent Arabic speakers came to existence. Even fluent speakers cannot always correctly determine the proper diacritization to use in certain sentences. There are many books, articles, magazines and letters that lack diacritization, which makes understanding their content problematic for most Arabic speakers. Moreover, manually adding diacritization to clarify the content is time consuming and can only be reliable through linguistics experts specializing in the Arabic language. Thus, the need for an automated diacritization system is eminent \cite{abandah2015automatic,belinkov2015arabic}.


Publishers, writers, producers and news agencies all care about delivering to their audiences easy to understand Arabic content. However, the cost to produce such content is high and time consuming.
In response to this problem, our work focuses on producing a powerful tool for automated Arabic text diacritization.
More formally, given a non-empty sequence of partially or non-diacritized Arabic text, find the correct diacritization for it. One of the aims of this work is to help in benchmarking efforts invested in this field. We also aim to show the high potential of Deep Learning (DL) approaches for this problem. In fact, we show that an existing DNN approach outperforms all existing tools and systems that are publicly available on our dataset.

DL approaches are producing ground breaking results in many tasks across different fields, such as Natural Language Processing (NLP), Computer Vision, Time-Series Analysis (TSA), etc. In fact, new and exciting applications of DL in NLP are among the reasons that the NLP market is expected to grow to tens of billions of dollars soon. Unfortunately, the use of DL within the Arabic NLP community is still limited even for NLP areas in which DL approaches are becoming dominant such as machine translation and sentiment analysis \cite{special_issue}.

The rest of this paper is organized as follows.
The next section presents a general literature survey of both research papers and existing tools for Arabic text diacritization.
The methodology we follow in benchmarking these tools and systems is discussed in Section~\ref{sec:method} including the benchmark dataset preparation steps and the different tools and systems/approaches under consideration. The results and discussion are given in Section~\ref{sec:res} before concluding the paper in Section~\ref{sec:conc}.

\section{Diacritization Systems and Approaches}
\label{sec:related}

There are two common approaches to address the Arabic text diacritization problem: rule-based approaches and machine learning (ML) approaches. The focus of this work is on the DL approach (a sub-field of ML) and we aim to show that it is superior to its publicly available competitors. This places it as the main benchmark system to beat by any future work.
The coverage given in this section is divided into two parts: DL-based (neural) approaches and baseline tools and systems.

\subsection{Neural Diacritization}

A recent survey on DL techniques for Arabic NLP tasks noted that limited attention has been paid to neural approaches for Arabic text diacritization \cite{dl4anlp}. On the other hand, there are some work published on non-neural approaches such as \cite{zitouni2009arabic,pasha2014madamira,shahrour2015improving,alnefaie2017automatic,bebah2014hybrid,chennoufi2017morphological,darwish2017arabic,fashwan2017shakkil,azmi2015survey}.
Such works are mainly based on linguistic rules and statistical treatments. For example, the MADAMIRA analyzer built by Pasha et al.~\cite{pasha2014madamira} provides diacritization, tokenization, part-of-speech tagging and other Arabic language processing tools, using morphological analysis.
Elshafei et al.~\cite{elshafei2006statistical} applied a statistical approach using hidden Markov model (HMM) to find the best diacritization format for words. This was done using Viterbi algorithm based on words' n-grams. They used Holy Quran as dataset.
Further discussion of the details of these approaches are outside the scope of this work. The only ones we discuss later are the ones with publicly available resources, which we use for comparison with the DNN approach.

Belinkov and Glass~\cite{belinkov2015arabic} presented a language-agnostic system for Arabic text diacritization. According to \cite{darwish2017arabic}, this is the only work on Arabic text diacritization that does not employ linguistic features and tools. The authors trained their system on diacritized text extracted from Arabic Treebank dataset without relying on additional resources. They used different types of neural networks in addition to letter embeddings to automatically diacritize Arabic text. The network types they considered are Feed-Forward, Long Short-Term Memory (LSTM), Bidirectional LSTM (B-LSTM) and stacked B-LSTM.
This approach is open-sourced and its results rival those of the state-of-the-art systems that rely on language-specific tools such as the MaxEnt approach of Zitouni and Sarikaya~\cite{zitouni2009arabic}.
%
A similar work done by Abandah et al.~\cite{abandah2015automatic} also used stacked B-LSTM on text extracted from Arabic Treebank, Tashkeela Corpus and the Holy Quran which achieved state-of-the-art performance after applying some language-related post-processing and error correction techniques.

\begin{figure}[h]
    \centering
    \includegraphics[width=0.5\textwidth]{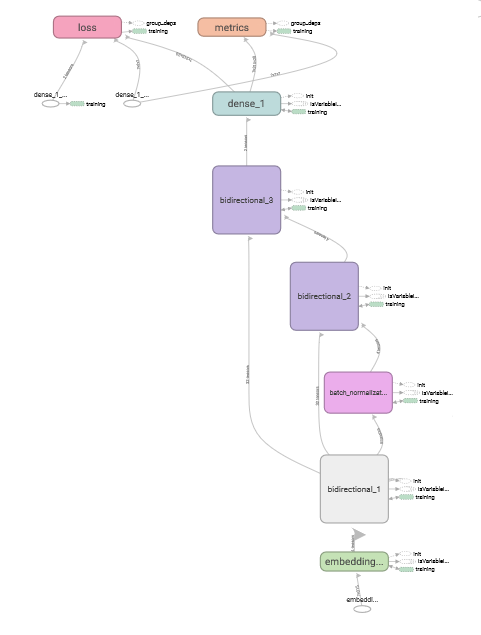}
    \caption{Shakkala model \cite{shakkala}.}
    \label{fig:shakkala}
\end{figure}

Finally and most importantly, the open-source project Shakkala was built by Barqawi and Zerrouki~\cite{shakkala} for Arabic text diacritization using B-LSTM networks in addition to character embeddings. The model is depicted in Figure~\ref{fig:shakkala}.
The model was trained on Tashkeela Corpus several times while removing the data with negative influence on the training process.
The system provides the diacritization service through an interactive web interface, without providing an API.\footnote{\url{https://ahmadai.com/shakkala}}
The website allows users to diacritize text containing up to 490 symbols. An error message is given if the input is longer.
The code is publicly available on Github\footnote{\url{https://github.com/Barqawiz/Shakkala}}. It has three different trained models. The first (and earliest) version is used in the website, while the third (and latest) version provides the best results but is limited to 315 characters at a time.

\subsection{Baseline Systems}
\label{sec:tools}

Searching through the Internet reveals few websites offering to automatically diacritize Arabic text. They are discussed in this subsection.

\softpara{Ali-Soft:}
Ali-Soft\footnote{\url{http://www.ali-soft.com}} provides simple diacritization and text to speech (TTS) services through interactive web interface with additional features to style the text while diacritizing it.
Ali-Soft allows the users to update the final diacritization results using predefined list of probable diacritizations.

After few test cases, one can easily notice that this system does not diacritize each letter in each word. In fact, most of the times, it does not diacritize the last letters. Moreover, it has some errors as shown in the following examples that prohibits us from including it in our experiments.
\begin{quote}
\centering
\<
المسلم من سلم المسلمون من لسانه ويده
> \\
Buckwalter Transliteration: Almslm mn slm Almslmwn mn lsAnh wydh \\
The sentence before diacritization. \\
\<
المُسْلِم مِن سُلَّم المُسْلِمُونَ مِن لِسانهُ وَيَدْهَ
> \\
Buckwalter Transliteration: Almusolim min sul$\sim$am Almusolimuwna min lisAnhu wayadoha \\
The sentence with Ali-Soft diacritization. \\
\<
اَلْمُسْلِمُ مَنْ سَلِمَ اَلْمُسْلِمُونَ مِنْ لِسَانِهِ وَيَدِهِ
> \\
Buckwalter Transliteration: Aalomusolimu mano salima Aalomusolimuwna mino lisaAnihi wayadihi \\
The correct diacritization. \\
\end{quote}
In addition to outputting incorrect diacritizations, Ali-Soft suffers from the following issues.

\begin{itemize}
\item It duplicates some letters as shown in the following examples.
\begin{center}
\begin{tabular}{|c|c|}
\hline
Input word & Output word \\ \hline
\<يؤذن> & 
\<يؤذنن> \\ \hline
\<ودونه> & 
\<ودوننه> \\ \hline
\<ثبت> & 
\<ثبتت> \\ \hline
\<رهنا> & 
\<رهننا> \\ \hline
\end{tabular}
\end{center}

\item Adding the Dagger Alif which is written as a short vertical stroke on top of an Arabic letter. It indicates a long \textipa{/a:/} sound where Alif is normally not written as shown in the following examples.
\begin{center}
\begin{tabular}{|c|c|}
\hline
Input word & Output word \\ \hline
\<الرحمن> & 
\<الرحمٰن> \\ \hline
\<الإلهية> & 
\<الإلٰهية> \\ \hline
\<هؤلاء> & 
\<هٰؤلاء> \\ \hline
\<اللهم> & 
\<اللٰهم> \\ \hline
\end{tabular}
\end{center}

\item Puts the Wasla sign on the letter (Alif) as shown in the following examples.
\begin{center}
\begin{tabular}{|c|c|}
\hline
Input word & Output word \\ \hline
\<ابن> & 
\<ٱبن> \\ \hline
\<امرأتان> & 
\<ٱمرأتان> \\ \hline
\<انقطع> & 
\<ٱنقطع> \\ \hline
\end{tabular}
\end{center}

\item It duplicates some words as shown in the following examples.
\begin{center}
\begin{tabular}{|c|c|}
\hline
Input word & Output word \\ \hline
\<الأنصارى> & 
\<الأنصاري الأنصاري> \\ \hline
\<يسارى> & 
\<يساري يساري يساري> \\ \hline
\<نفسى> & 
\<نفسي نفسي نفسي نفسي> \\ \hline
\<أى> & 
\<أي أي أي أي أي أي> \\ \hline
\end{tabular}
\end{center}
\end{itemize}

\softpara{Farasa:}
Farasa\footnote{\url{http://alt.qcri.org/farasa}} provides multiple Arabic NLP services and one of them is partial diacritization. These services are available through a web interface, an API and a standalone executable code.\footnote{\url{http://qatsdemo.cloudapp.net/farasa}}
However, like Ali-Soft, the diacritization system of Farasa has a few issues as shown in the following examples. These issues are more manageable compared with the ones suffered by Ali-Soft and we discuss in Section~\ref{sec:systems_issues} how to handle them before including Farasa in our comparison.
\begin{center}
\begin{tabular}{|c|c|}
\hline
Input word & Output word \\ \hline
\<لِلَّهِ> & 
\<لِاللَّهِ> \\ \hline
\<لِلَّذِي> & 
\<لِالَّذي> \\ \hline
\<مِنَّةٌ> & 
\<مِنَّةْهُ> \\ \hline
\end{tabular}
\end{center}

\softpara{Harakat:}
Harakat\footnote{\url{https://harakat.ae}} provides diacritization service through an interactive web interface and an API.\footnote{\url{https://multillect.com/en/apidoc}}
The interactive web interface allows users to diacritize text containing about 650 symbols (giving an error message for a longer text) and it takes about 3 to 5 seconds to diacritize them, while the API diacritizes 1M symbols for \$5.
Like Farasa, the Harakat system has a variety of issues as shown in the following examples that must be handled before including it in our experiments.
\begin{quote}
\centering
\<
ذهب علي كثير
> \\
Buckwalter Transliteration: *hb Ely kvyr \\
The sentence before diacritization. \\
\<
ذَهَبَ عَلَى كَثِيرٍ
> \\
Buckwalter Transliteration: *ahaba EalaY kaviyrK \\
The sentence with Harakat diacritization. \\
\<
ذَهَبُ عَلِيٍّ كَثِيرٌ
> \\
Buckwalter Transliteration: *ahabu Ealiy$\sim$K kaviyrN \\
The correct diacritization. \\
\end{quote}
In addition to the incorrect diacritization, note that Harakat changed the last letter from
\<
ي
>
to
\<
ى
>
in
\<
علي
>
(the name `Ali') which is changed to
\<
على
>
(the preposition `on' in English).
Harakat has other issues that affects comparison with other systems such as modifying the input words as shown in the following examples.

\begin{center}
\begin{tabular}{|c|c|}
\hline
Input word & Output word \\ \hline
\<فضل> & 
\<فضلا> \\ \hline
\<بين> & 
\<بي> \\ \hline
\<عجرة> & 
\<جرة> \\ \hline
\end{tabular}
\end{center}

\softpara{MADAMIRA:}
MADAMIRA\footnote{\url{https://camel.abudhabi.nyu.edu/madamira}} is an Arabic analyzer that provides features like POS tagging, tokenization, and diacritization. The system is built for either MSA or Egyptian Dialect. Both release types (bundled and unbundled) of MADAMIRA come packaged with a morphological database called Aramorph. However, only the bundled one comes with SAMA database, that according to the authors, provides a better prediction accuracy. The system has a couple of issues including changing symbols from input to Arabic characters
(for example, changing
`\}'
to `\<ئ>')
and adding the Dagger Alif (mentioned previously) to certain words.

\softpara{Mishkal}
Mishkal\footnote{\url{https://tahadz.com/mishkal}} is an open source rule-based Arabic text diacritization system. The system also provides some analytical information about the input words. It is provided through a web interface, desktop application, and command-line interface. The system has other issues that affects comparison with other systems such as modifying the input words as shown in the following examples.

\begin{center}
\begin{tabular}{|c|c|}
\hline
Input word & Output word \\ \hline
\<وعشرون> & 
\<عشرون> \\ \hline
\<وثمانية> & 
\<ثمانية> \\ \hline
\<كلمة :> & 
\<كلمة:> \\ \hline
\end{tabular}
\end{center}

\softpara{Tashkeela-Model}
Tashkeela-Model\footnote{\url{ https://github.com/Anwarvic/Tashkeela-Model}} is an n-gram model for Arabic language diacritization trained using the Tashkeela dataset. The model is not available through an interactive web interface or an API, but it is publicly available on GitHub.


\section{Methodology}
\label{sec:method}

This section gives a fully detailed description of what dataset was used and how it was cleaned and prepared for training and testing. Then, it presents the different diacritization approaches tested on this dataset.

\subsection{Dataset}

The training, validation and testing datasets come from the same distribution, extracted from the Tashkeela Corpus\footnote{\url{https://sourceforge.net/projects/tashkeela}} and the simplified version of Holy Quran. Tashkeela consists of 97 CA books and 293 MSA files compiled from books, articles, news, speeches and school lessons.
Basic statistics about the conetent of the dataset and the usage of diacritics within are given in Tables~\ref{tab4_0} and \ref{tab4_1}.

\begin{table}
\centering
\caption{Dataset statistics}
\label{tab4_0}
\begin{tabular}{|c|c|c|}
\hline
 & CA and Holy Quran & MSA \\ \hline
Words Count & 66,547K & 801K \\ \hline
Lines Count & 1,706K & 49K \\ \hline
Avg Chars/Word & 4.05 & 4.41 \\ \hline
Avg Words/Line & 38.98 & 16.06 \\ \hline
\end{tabular}
\end{table}

\begin{table}
\centering
\caption{Diacritics usage in the dataset}
\label{tab4_1}
\begin{tabular}{|c|c|c|}
\hline
 & CA and Holy Quran & MSA \\ \hline
No Diacritics (\%) & 21.79 & 40.25 \\ \hline
One Diacritic (\%) & 72.89 & 55.29 \\ \hline
Two Diacritics (\%) & 5.30 & 4.27 \\ \hline
Error Diacritics (\%) & 0.002 & 0.17 \\ \hline
\end{tabular}
\end{table}

\subsubsection{Tashkeela Corpus Issues}
\label{sec:tashkeela_issues}

\begin{figure}
    \centering
    \includegraphics[width=0.5\textwidth]{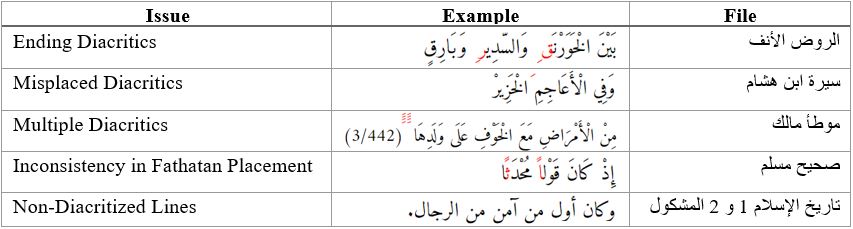}
    \caption{Examples of the Tashkeela corpus issues.}
    \label{issues}
\end{figure}

During the cleaning process we encountered various issues with the diacritization in the Tashkeela Corpus, as listed below:
\begin{itemize}
\item Ending Diacritics: The diacritic belongs to the end of the word, but is separated from it with one or more whitespaces. An example is shown in Figure~\ref{issues}.
    
\item Misplaced Diacritics: Diacritics following a non-Arabic character such as a whitespace, numbers and punctuations. An example is shown in Figure~\ref{issues}.

\item Multiple Diacritics: Multiple diacritics appear on a single character. An example is shown in Figure~\ref{issues}.

\item Inconsistency in Fathatan Placement: In most cases, for diacritizing a character with Fathatan diacritic, an extra character Alif
`\<
ا
>'
must be added after the character. There are two schools for placing the diacritic, one which puts it after the Alif, and the other one puts it before the Alif (on the character itself). Both might appear in the same sentence. An example is shown in Figure~\ref{issues}.

\item Non-Diacritized Lines: Some books of CA and many files in MSA contain lines without any diacritization. An example is shown in Figure~\ref{issues}.

\item Non-Diacritized Files: Some MSA files do not have any diacritization at all; e.g.,
\linebreak
`\<
عن
النعمة
>'.

\item
Newline characters do not separate paragraphs correctly, instead, they can occur in the middle of a paragraph.
Examples can be found in
'\<
بريقة محمودية في شرح طريقة محمدية وشريعة نبوية
>'
file from CA.

\end{itemize}

\subsubsection{Cleaning Process}
The cleaning process is divided into several steps as follows:
\begin{itemize}
\item Removing HTML tags: A minority of files use HTML format. BeautifulSoup\footnote{\url{https://www.crummy.com/software/BeautifulSoup}} library is used to strip the HTML tags from the text.

\item Remove URLs: Using regex to remove URLs from the content of the files.

\item Fix Diacritization Issue: Solves the first four mentioned issues from Section~\ref{sec:tashkeela_issues} using regex by removing the whitespaces for `Ending Diacritics,' removing `Misplaced Diacritics,' keeping only the first diacritic in `Multiple Diacritics' and change all after-Alif cases (0.67\%) to before-Alif cases (99.33\%).

\item Remove English Letters: Remove English letters that are scattered throughout the files in the form of incomplete URLs or HTML/CSS code leftovers.

\item Remove the special Arabic character (Kashida): This character is removed from the text because it only has a calligraphic effect that does not change the word diacritization.

\item Separate Numbers: Whitespaces are added before and after each number using regex, because many of them are stuck to other words.


\item Remove Multiple Whitespaces: Using regex, consecutive whitespaces are reduced to a single space character if not containing a new-line characters, a new-line character otherwise.

\end{itemize}

Finally, we create a non-diacritized version of the dataset. This gives us a parallel corpus of diacritized and non-diacritized sentences.

\subsubsection{Dataset Split}

About 55K lines are randomly chosen from the CA and HQ dataset only. The lines are extracted such that the diacritics to Arabic characters rate is greater than 80\%. Moreover, 5K lines from them are randomly chosen and split evenly between the validation and testing datasets. Statistics about the dataset size and diacritics usage (after the cleaning process) are given in Tables~\ref{tab4_2a} and \ref{tab4_2b}.

\begin{table}
\centering
\caption{Dataset statistics after the cleaning process}
\label{tab4_2a}
\begin{tabular}{|c|c|c|c|}
\hline
 & Training & Validation & Testing \\ \hline
Words Count & 2,103K & 102K & 107K \\ \hline
Lines Count & 50K & 2.5K & 2.5K \\ \hline
Chars per Word (AVG) & 3.97 & 3.97 & 3.97 \\ \hline
Words per Line (AVG) & 42.06 & 40.97 & 42.89 \\ \hline
\end{tabular}
\end{table}

\begin{table}
\centering
\caption{Diacritics usage in the cleaned dataset}
\label{tab4_2b}
\begin{tabular}{|c|c|c|c|}
\hline
 & Training & Validation & Testing \\ \hline
No Diacritics (\%) & 17.78 & 17.75 & 17.80 \\ \hline
One Diacritic (\%) & 77.17 & 77.19 & 77.22 \\ \hline
Two Diacritics (\%) & 5.03 & 5.05 & 4.97 \\ \hline
Error Diacritics (\%) & 0 & 0 & 0  \\ \hline
\end{tabular}
\end{table}

In order to enhance the reproducibility of our work, the cleaned dataset is publicly available on GitHub along with all code related to the cleaning process and comparison procedure.\footnote{\url{https://github.com/AliOsm/arabic-text-diacritization}}
By publicly providing access to our dataset and resources, our goal is to provide the first freely available benchmark dataset for Arabic text diacritization.

\subsection{Evaluation Metrics}

The comparison process was based on the following two metrics.
\begin{itemize}
\item Diacritic Error Rate (DER): The percentage of misclassified Arabic characters whether the character has 0, 1 or 2 diacritics.

\item Word Error Rate (WER): The percentage of Arabic words which have at least one misclassified Arabic character.
\end{itemize}
Note that other works, such as \cite{zitouni2009arabic}, use DER/WER definitions which take into account all numbers and punctuations; this results in a decreased DER/WER compared to our definition since numbers and punctuations do not affect the error rates.
The following example clarifies the difference between the two definitions.
\begin{quote}
\centering
\<
انْظُرْ إِلَى السَّمَاءِ ، كَمْ هِيَ جَمِيلَةٌ !!!
> \\
Buckwalter Transliteration: AnoZuro $<$ilaY Als$\sim$amaA'i, kamo hiya jamiylapN !!! \\
Translation: Look at the sky, how beautiful it is!!!.
\end{quote}
Assuming that the system misclassifies the character
`\<ظ>'
with Fatha instead of Damma, then the DER/WER will be as shown in Tables~\ref{tab:newder} and \ref{tab:newwer}.

\begin{table*}
\centering
\caption{Comparison of our definition of DER and the definition of Zitouni and Sarikaya~\cite{zitouni2009arabic}}
\label{tab:newder}
\begin{tabular}{|c|c|c|c|c|}
\hline
\multirow{2}{*}{DER} & With case ending & Without case ending & With case ending & Without case ending \\ \cline{2-5} 
 & \multicolumn{2}{c|}{Including 'no diacritic'} & \multicolumn{2}{c|}{Excluding 'no diacritic'} \\ \hline
Ours
& 4.55\% & 6.25\% & 6.25\% & 9.09\% \\ \hline
Zitouni and Sarikaya~\cite{zitouni2009arabic} & 3.85\% & 5.56\% & 6.25\% & 9.09\% \\ \hline
\end{tabular}
\end{table*}

\begin{table*}
\centering
\caption{Comparison of our definition of WER and the definition of Zitouni and Sarikaya~\cite{zitouni2009arabic}}
\label{tab:newwer}
\begin{tabular}{|c|c|c|c|c|}
\hline
\multirow{2}{*}{WER} & With case ending & Without case ending & With case ending & Without case ending \\ \cline{2-5} 
 & \multicolumn{2}{c|}{Including 'no diacritic'} & \multicolumn{2}{c|}{Excluding 'no diacritic'} \\ \hline
Ours
& 16.67\% & 16.67\% & 16.67\% & 16.67\% \\ \hline
Zitouni and Sarikaya~\cite{zitouni2009arabic} & 12.50\% & 12.50\% & 12.50\% & 12.50\% \\ \hline
\end{tabular}
\end{table*}

Both DER and WER can be calculated with/without case ending, and with/without the `no diacritic' class \cite{zitouni2009arabic}.
\begin{itemize}
\item With/Without case ending: Determines whether to take the last letter diacritic into account or not. This is used because diacritizing the last letter is considered to be a harder problem compared to diacritizing other letters.

\item With/Without `no diacritic' class: Determines whether to take letters with no diacritic in the original data into account or not. This is used because some systems provide full diacritization and others provide partial diacritization.
\end{itemize}

\subsection{Systems Compared}
\label{sec:systems_issues}

The goal of this work is to show the superiority of the neural approach compared with other approaches. The neural approach we consider is Shakkala of Barqawi and Zerrouki~\cite{shakkala}. The reason for selecting this system is because the Shakkala code and the resources used to build it are publicly available.
As for the non-neural approaches, most of the ones mentioned in Section~\ref{sec:tools} are considered in our experiments. The following points are taken into account during the testing procedure.
\begin{itemize}
\item For all systems, after-Alif cases are changed to before-Alif during DER and WER calculation process to match the original dataset diacritization standard.

\item Ali-Soft testing was omitted due to the excessive amount of issues.

\item Farasa, MADAMIRA and Mishkal testing process had some issues which were manually fixed.

\item Harakat testing process skipped all 143 lines which had issues.

\item Shakkala testing data was split into lines of lengths not exceeding 315 characters because their third model was used in testing.
\end{itemize}

\section{Results and Discussion}
\label{sec:res}

In this section, we present the results of our experiments and discuss them.
The comparison results are provided in Tables~\ref{tab:der} and \ref{tab:wer} (best results are shown in bold).

\begin{table*}
\centering
\caption{DER results}
\label{tab:der}
\begin{tabular}{|c|c|c|c|c|}
\hline
\multirow{2}{*}{DER} & With case ending & Without case ending & With case ending & Without case ending \\ \cline{2-5} 
 & \multicolumn{2}{c|}{Including 'no diacritic'} & \multicolumn{2}{c|}{Excluding 'no diacritic'} \\ \hline
Farasa & 21.43\% & 23.93\% & 24.90\% & 27.55\% \\ \hline
Harakat & 18.37\% & 17.03\% & 20.64\% & 18.55\% \\ \hline
MADAMIRA (Aramorph) & 34.38\% & 29.94\% & 40.03\% & 33.87\% \\ \hline
Mishkal & 16.09\% & 13.78\% & 17.59\% & 14.22\% \\ \hline
Tashkeela-Model & 49.96\% & 52.96\% & 58.50\% & 60.92\% \\ \hline
Shakkala & \textbf{3.73\%} & \textbf{2.88\%} & \textbf{4.36\%} & \textbf{3.33\%} \\ \hline
\end{tabular}
\end{table*}

\begin{table*}
\centering
\caption{WER results}
\label{tab:wer}
\begin{tabular}{|c|c|c|c|c|}
\hline
\multirow{2}{*}{WER} & With case ending & Without case ending & With case ending & Without case ending \\ \cline{2-5} 
 & \multicolumn{2}{c|}{Including 'no diacritic'} & \multicolumn{2}{c|}{Excluding 'no diacritic'} \\ \hline
Farasa & 58.88\% & 53.13\% & 57.28\% & 51.84\% \\ \hline
Harakat & 41.83\% & 32.03\% & 38.33\% & 28.29\% \\ \hline
MADAMIRA (Aramorph) & 76.58\% & 59.07\% & 75.39\% & 57.22\% \\ \hline
Mishkal & 39.78\% & 26.42\% & 35.63\% & 21.92\% \\ \hline
Tashkeela-Model & 96.80\% & 94.16\% & 96.03\% & 92.45\% \\ \hline
Shakkala & \textbf{11.19\%} & \textbf{6.53\%} & \textbf{10.89\%} & \textbf{6.37\%} \\ \hline
\end{tabular}
\end{table*}


The results clearly show that the DL approach (Shakkala) is significantly better than all other approaches in terms of both DER and WER. Mishkal and Harakat are the best systems among the non-neural ones, but their performance in not comparable to Shakkala. Another advantage of Shakkala is its remarkable ability to handle the difficult case of diacritizing the last letter of each word.

\section{Conclusion}
\label{sec:conc}

According to many researchers such as Taha Zerrouki, diacritizing Arabic text is among the most challenging problems in Arabic NLP. In order to move towards providing effective solutions to this problem, open-source resources are needed. In this work, we present the first free-for-all cleaned benchmark dataset for this problem. Extracted from the Tashkeela Corpus, the dataset consists of 55K lines containing about 2.3M words. Moreover, we provide a critical review for the currently existing systems and tools for Arabic text diacritization and perform an empirical study to compare the performance of six of them on our dataset. Moreover, we revise the definitions of the most common accuracy measures used for Arabic text diacritization, viz., Diacritic Error Rate (DER) and Word Error Rate (WER). The revised measures do not take into account numbers and punctuations, which make them more strict. The results of our experiments show that the neural Shakkala system significantly outperforms traditional rule-based approaches and other closed-source tools with DER and WER values as low as 2.88\% and 6.37\%, respectively, compared with the lowest DER and WER values for a non-neural system (Mishkal) which are 13.78\% and 21.92\%, respectively.

\bibliographystyle{IEEEtran}
\bibliography{main}

\begin{thebibliography}{10}
\providecommand{\url}[1]{#1}
\csname url@samestyle\endcsname
\providecommand{\newblock}{\relax}
\providecommand{\bibinfo}[2]{#2}
\providecommand{\BIBentrySTDinterwordspacing}{\spaceskip=0pt\relax}
\providecommand{\BIBentryALTinterwordstretchfactor}{4}
\providecommand{\BIBentryALTinterwordspacing}{\spaceskip=\fontdimen2\font plus
\BIBentryALTinterwordstretchfactor\fontdimen3\font minus
  \fontdimen4\font\relax}
\providecommand{\BIBforeignlanguage}[2]{{%
\expandafter\ifx\csname l@#1\endcsname\relax
\typeout{** WARNING: IEEEtran.bst: No hyphenation pattern has been}%
\typeout{** loaded for the language `#1'. Using the pattern for}%
\typeout{** the default language instead.}%
\else
\language=\csname l@#1\endcsname
\fi
#2}}
\providecommand{\BIBdecl}{\relax}
\BIBdecl

\bibitem{habash2010introduction}
N.~Y. Habash, ``Introduction to arabic natural language processing,''
  \emph{Synthesis Lectures on Human Language Technologies}, vol.~3, no.~1, pp.
  1--187, 2010.

\bibitem{farghaly2009arabic}
A.~Farghaly and K.~Shaalan, ``Arabic natural language processing: Challenges
  and solutions,'' \emph{ACM Transactions on Asian Language Information
  Processing (TALIP)}, vol.~8, no.~4, p.~14, 2009.

\bibitem{asa_survey}
M.~Al-Ayyoub, A.~A. Khamaiseh, Y.~Jararweh, and M.~N. Al-Kabi, ``A
  comprehensive survey of arabic sentiment analysis,'' \emph{Information
  Processing \& Management}, vol.~56, no.~2, pp. 320--342, 2019.

\bibitem{abandah2015automatic}
G.~A. Abandah, A.~Graves, B.~Al-Shagoor, A.~Arabiyat, F.~Jamour, and
  M.~Al-Taee, ``Automatic diacritization of arabic text using recurrent neural
  networks,'' \emph{International Journal on Document Analysis and Recognition
  (IJDAR)}, vol.~18, no.~2, pp. 183--197, 2015.

\bibitem{belinkov2015arabic}
Y.~Belinkov and J.~Glass, ``Arabic diacritization with recurrent neural
  networks,'' in \emph{Proceedings of the 2015 Conference on Empirical Methods
  in Natural Language Processing}, 2015, pp. 2281--2285.

\bibitem{special_issue}
Y.~Jararweh, M.~Al-Ayyoub, and E.~Benkhelifa, ``Advanced arabic natural
  language processing (anlp) and its applications: Introduction to the special
  issue,'' \emph{Information Processing \& Management}, vol.~56, no.~2, pp.
  259--261, 2019.

\bibitem{dl4anlp}
M.~Al-Ayyoub, A.~Nuseir, K.~Alsmearat, Y.~Jararweh, and B.~Gupta, ``Deep
  learning for arabic nlp: A survey,'' \emph{Journal of computational science},
  vol.~26, pp. 522--531, 2018.

\bibitem{zitouni2009arabic}
I.~Zitouni and R.~Sarikaya, ``Arabic diacritic restoration approach based on
  maximum entropy models,'' \emph{Computer Speech \& Language}, vol.~23, no.~3,
  pp. 257--276, 2009.

\bibitem{pasha2014madamira}
A.~Pasha, M.~Al-Badrashiny, M.~T. Diab, A.~El~Kholy, R.~Eskander, N.~Habash,
  M.~Pooleery, O.~Rambow, and R.~Roth, ``Madamira: A fast, comprehensive tool
  for morphological analysis and disambiguation of arabic.'' in \emph{LREC},
  vol.~14, 2014, pp. 1094--1101.

\bibitem{shahrour2015improving}
A.~Shahrour, S.~Khalifa, and N.~Habash, ``Improving arabic diacritization
  through syntactic analysis,'' in \emph{Proceedings of the 2015 Conference on
  Empirical Methods in Natural Language Processing}, 2015, pp. 1309--1315.

\bibitem{alnefaie2017automatic}
R.~Alnefaie and A.~M. Azmi, ``Automatic minimal diacritization of arabic
  texts,'' \emph{Procedia Computer Science}, vol. 117, pp. 169--174, 2017.

\bibitem{bebah2014hybrid}
M.~Bebah, C.~Amine, M.~Azzeddine, and L.~Abdelhak, ``Hybrid approaches for
  automatic vowelization of arabic texts,'' \emph{arXiv preprint
  arXiv:1410.2646}, 2014.

\bibitem{chennoufi2017morphological}
A.~Chennoufi and A.~Mazroui, ``Morphological, syntactic and diacritics rules
  for automatic diacritization of arabic sentences,'' \emph{Journal of King
  Saud University-Computer and Information Sciences}, vol.~29, no.~2, pp.
  156--163, 2017.

\bibitem{darwish2017arabic}
K.~Darwish, H.~Mubarak, and A.~Abdelali, ``Arabic diacritization: Stats, rules,
  and hacks,'' in \emph{Proceedings of the Third Arabic Natural Language
  Processing Workshop}, 2017, pp. 9--17.

\bibitem{fashwan2017shakkil}
A.~Fashwan and S.~Alansary, ``Shakkil: an automatic diacritization system for
  modern standard arabic texts,'' in \emph{Proceedings of the Third Arabic
  Natural Language Processing Workshop}, 2017, pp. 84--93.

\bibitem{azmi2015survey}
A.~M. Azmi and R.~S. Almajed, ``A survey of automatic arabic diacritization
  techniques,'' \emph{Natural Language Engineering}, vol.~21, no.~3, pp.
  477--495, 2015.

\bibitem{elshafei2006statistical}
M.~Elshafei, H.~Al-Muhtaseb, and M.~Alghamdi, ``Statistical methods for
  automatic diacritization of arabic text,'' in \emph{The Saudi 18th National
  Computer Conference. Riyadh}, vol.~18, 2006, pp. 301--306.

\bibitem{shakkala}
\BIBentryALTinterwordspacing
A.~Barqawi and T.~Zerrouki, ``Shakkala, arabic text vocalization,'' 2017.
  [Online]. Available: \url{https://github.com/Barqawiz/Shakkala}
\BIBentrySTDinterwordspacing

\end{thebibliography}
\balance

\end{document}